\documentclass[lettersize,journal]{IEEEtran}
\usepackage{amsmath,amsfonts}
\usepackage{algorithmic}
\usepackage{algorithm}
\usepackage{array}

\usepackage{textcomp}
\usepackage{stfloats}
\usepackage{url}
\usepackage{verbatim}
\usepackage{graphicx}
\usepackage{cite}
\usepackage{enumitem}
\usepackage{multirow}
\usepackage{multicol}
\usepackage{rotating}
\usepackage[table,xcdraw]{xcolor}
\usepackage{xcolor}
\usepackage{subfigure}

\usepackage{amssymb}
\usepackage{booktabs}
\usepackage{times}
\usepackage{epsfig}
\usepackage{bbm}
\usepackage{pifont}
\usepackage{hyperref}
\hypersetup{hidelinks,
	colorlinks=true,
	allcolors=black,
	pdfstartview=Fit,
	breaklinks=true}
\usepackage{float}

\begin{document}

\title{Learnable Differencing Center for Nighttime \\ Depth Perception}


\author{Zhiqiang Yan, Yupeng Zheng, Chongyi Li,~\IEEEmembership{Senior Member,~IEEE,} Jun Li, and Jian Yang,~\IEEEmembership{Member,~IEEE}
\thanks{
Zhiqiang Yan, Jun Li, and Jian Yang are with Nanjing University of Science and Technology, China ({yanzq,junli,csjyang}@njust.edu.cn). 
Yupeng Zheng is with Chinese Academy of Sciences, China (zhengyupeng2022@ia.ac.cn). 
Chongyi Li is with Nankai University, China (lichongyi25@gmail.com).
}
}

\markboth{Journal of \LaTeX\ Class Files,~Vol.~14, No.~8, August~2021}%
{Shell \MakeLowercase{\textit{et al.}}: A Sample Article Using IEEEtran.cls for IEEE Journals}


\maketitle

\begin{abstract}
Depth completion is the task of recovering dense depth maps from sparse ones, usually with the help of color images. Existing image-guided methods perform well on daytime depth perception self-driving benchmarks, but struggle in nighttime scenarios with \emph{poor visibility} and \emph{complex illumination}. To address these challenges, we propose a simple yet effective framework called LDCNet. Our key idea is to use Recurrent Inter-Convolution Differencing (RICD) and Illumination-Affinitive Intra-Convolution Differencing (IAICD) to enhance the nighttime color images and reduce the negative effects of the varying illumination, respectively. RICD explicitly estimates global illumination by differencing two convolutions with different kernels, treating the small-kernel-convolution feature as the center of the large-kernel-convolution feature in a new perspective. IAICD softly alleviates local relative light intensity by differencing a single convolution, where the center is dynamically aggregated based on neighboring pixels and the estimated illumination map in RICD. On both nighttime depth completion and depth estimation tasks, extensive experiments demonstrate the effectiveness of our LDCNet, reaching the state of the art. 
\end{abstract}


\section{Introduction}\label{introduction}
\begin{figure}[t]
\centering
\subfigure[RGB]{
\includegraphics[width=0.241\columnwidth]{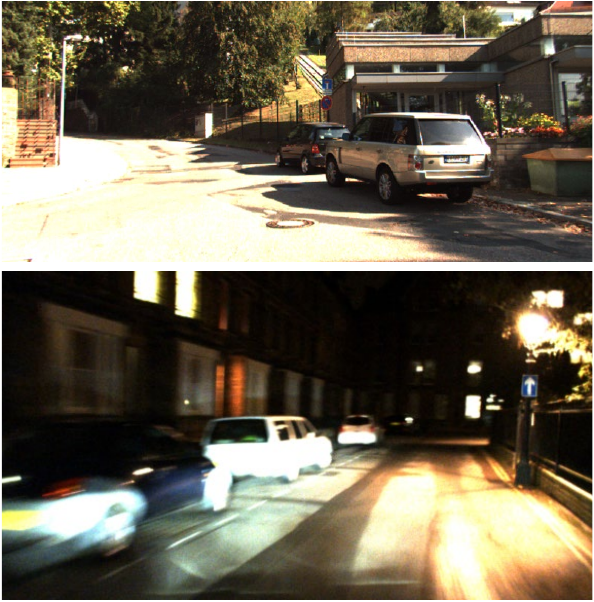}
}
\hspace{-13.5pt}
\subfigure[NLSPN \cite{park2020nonlocal}]{
\includegraphics[width=0.241\columnwidth]{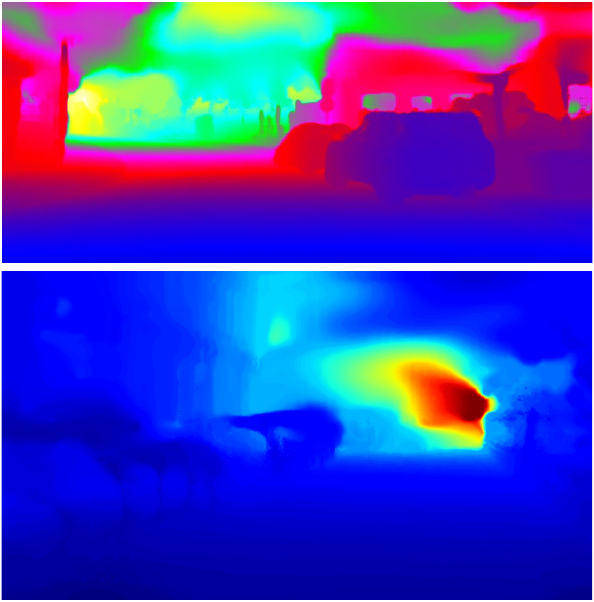}
}
\hspace{-13.5pt}
\subfigure[RigNet \cite{yan2022rignet}]{
\includegraphics[width=0.241\columnwidth]{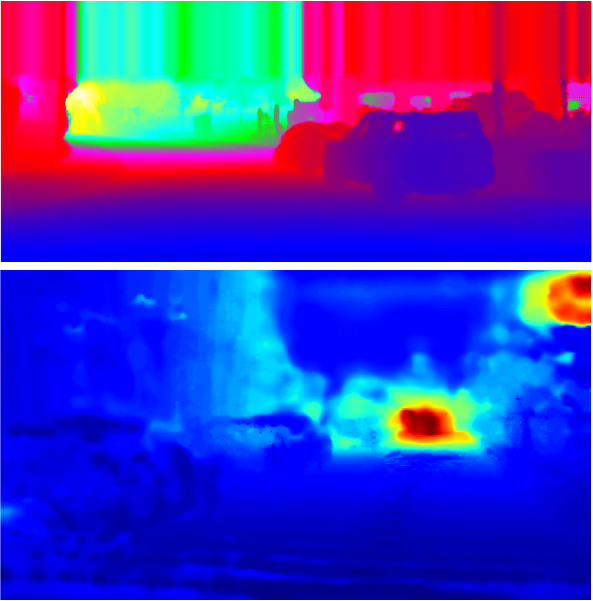}
}
\hspace{-13.5pt}
\subfigure[CFormer \cite{Zhang2023CompletionFormer}]{
\includegraphics[width=0.241\columnwidth]{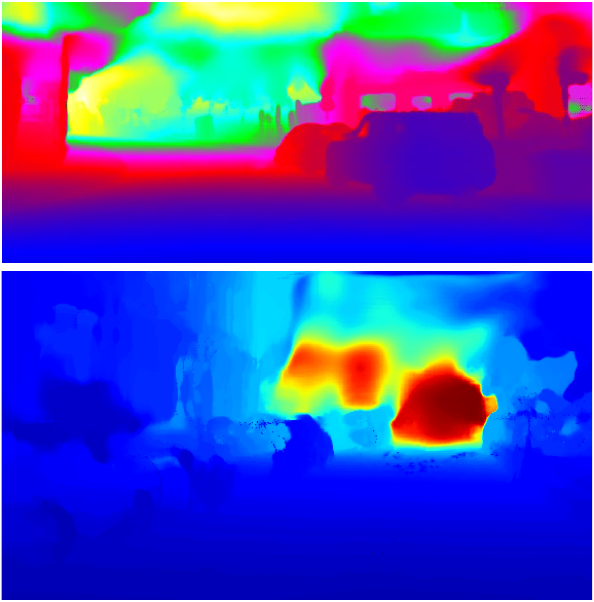}
}
\vspace{-5pt}
\caption{Visual results of different image-guided depth completion methods in daytime (first row) and nighttime (second row) scenarios.}\label{fig_day_night_comparison}
\vspace{-9pt}
\label{channel_comparison}
\end{figure}

\IEEEPARstart{D}epth completion \cite{hu2022deep} aims to predict dense depth maps from sparse ones and the corresponding color images. It is an essential task in computer vision and has been widely used in various applications, such as augmented reality \cite{dey2012tablet,ma2018self}, 3D scene reconstruction \cite{park2020nonlocal,yan2022learning}, and self-driving \cite{yan2022desnet,Zhang2023CompletionFormer}. In the past few years, plenty of image-guided methods \cite{liu2021fcfr,yan2022rignet,lin2022dynamic,Zhang2023CompletionFormer} have been proposed for depth completion under daytime conditions, \emph{e.g.}, the well-known KITTI benchmark \cite{Uhrig2017THREEDV}. However, there are very few approaches focusing on the more challenging nighttime scenarios. As we know that nighttime depth-aware self-driving is especially important but difficult. As shown in Fig.\ref{fig_day_night_comparison}, existing state-of-the-art image-guided depth completion methods \cite{park2020nonlocal,yan2022rignet,Zhang2023CompletionFormer} perform well in daytime conditions but struggle in challenging nighttime scenarios. This is because the sparse depths from LiDAR are illumination-invariant while the color images are highly affected by visibility and illumination variations. Therefore, we identify the key challenge of nighttime depth completion as the guidance from color images, which heavily suffer from \emph{poor visibility} and \emph{complex illumination}. 

\begin{figure}[t]
 \centering
 \includegraphics[width=0.86\columnwidth]{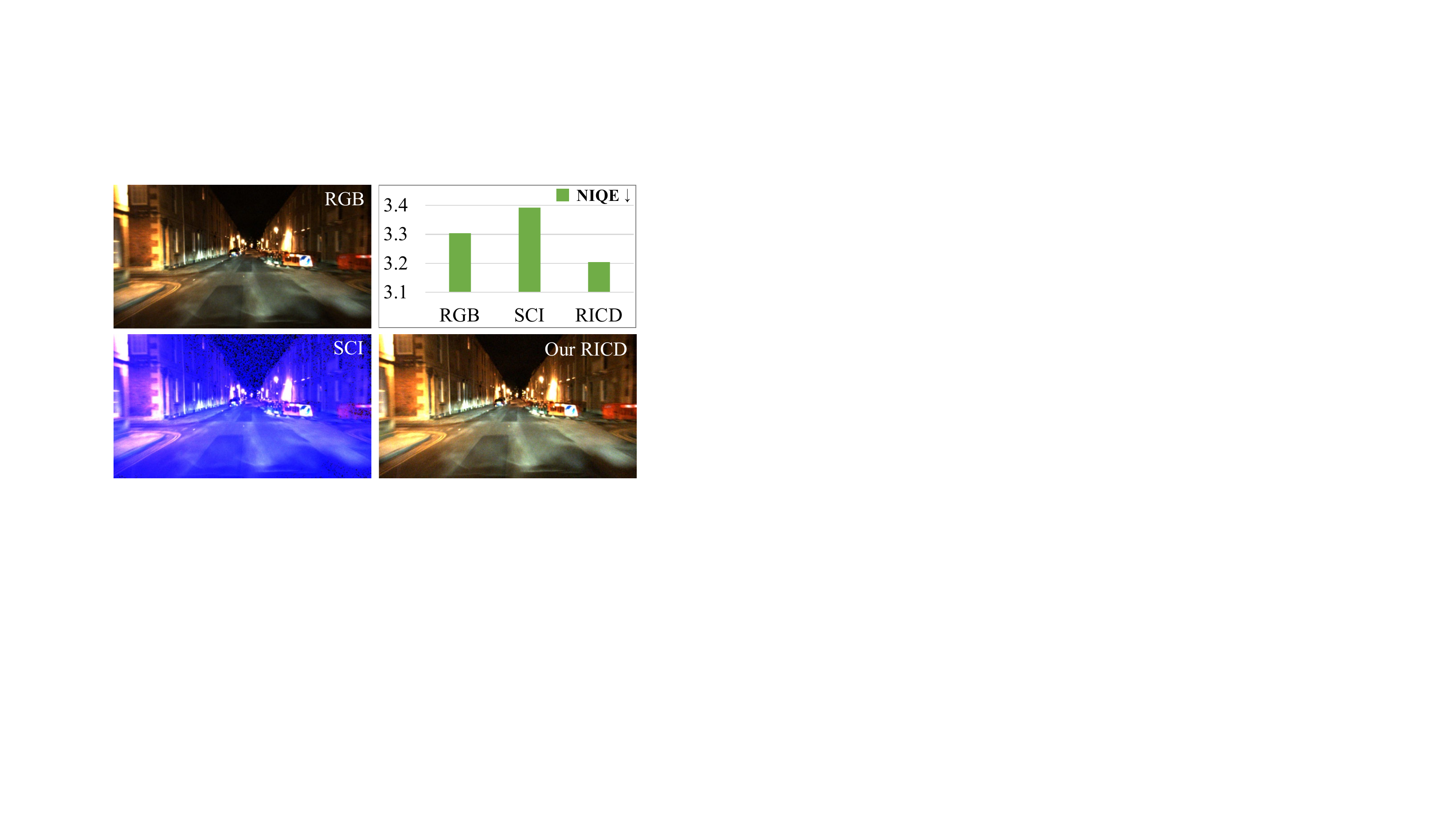}
 \vspace{-7pt}
 \caption{Visual comparison of SCI \cite{ma2022toward} and our RICD enhancement. NIQE: Natural Image Quality Evaluator \cite{wang2013naturalness}, a completely blind no-reference metric.}
 \label{fig_color_cast}
 \vspace{-7pt}
\end{figure}

\emph{For poor visibility}: A possible solution is to leverage existing low-light image enhancement techniques \cite{guo2020zero,ma2022toward,zhang2023fast} to improve the visibility of nighttime color images. Since there are no paired clear images available as supervisory signals, self-supervised methods \cite{guo2020zero,li2021learning,ma2022toward} for nighttime depth perception are preferred. However, we find that they cannot generate very reasonable illumination maps, resulting in extremely untrustworthy enhanced color images for safe self-driving. For example, Fig.\ref{fig_color_cast} shows that the state-of-the-art model \cite{ma2022toward} suffers from serious color cast. To tackle this issue, we propose \emph{Recurrent Inter-Convolution Differencing} (RICD), which explicitly and gradually estimates global illumination to improve the poor visibility, by using continuous differencing between two convolutions with different kernels. It is a new perspective to treat the small-kernel-convolution feature as the center of the large-kernel-convolution feature. Moreover, we recognize that convolution subtraction \cite{shi2022symmetric} is useful for modeling uncertainty \cite{kendall2017uncertainties} where the target pixels are difficult to predict accurately. In the nighttime image above, we can easily observe that there are many areas with underexposure, overexposure, and terminatorn (the junction area between light and dark) effects due to varying illumination, resulting in more and higher uncertainty than usual. Based on these priors, we transform the uncertainty in nighttime scenarios into relative light intensity by applying continuous convolution differencing. Such differencing features that capture explicit light intensity information are essential for predicting valid illumination. Consequently, RICD contributes to robust visibility enhancement of nighttime color images with more naturalistic visual effect in Fig \ref{fig_color_cast}.

\emph{For complex illumination}: However, even after applying RICD enhancement, the distribution of relative light intensity in nighttime color images is still much more complex than in daytime conditions. For instance, there are lots of terminatorn areas with varying illumination, which are difficult for standard convolutions to handle. Fortunately, inspired by the Local Binary Pattern operator \cite{boulkenafet2015face} that is robust to illumination variation, a series of central convolution differencing algorithms \cite{yu2020searching,yu2020fas,su2021pixel,tan2022semantic} are devised to address these challenging scenarios. Nevertheless, their differencing centers are typically fixed, leading to restricted applicability, especially for self-driving where safety is incredibly important. For example, when the center contains noise or lies on terminatorn, these algorithms would additionally introduce negative reference information, thus resulting in unsatisfactory illumination robustness. To tackle this issue, we propose \emph{Illumination-Affinitive Intra-Convolution Differencing} (IAICD) that learns reasonable differencing center within a single convolution. For one thing, IAICD can reduce the latent impact of noise and predict an adaptive differencing center based on the surrounding neighbors. For another, the estimated illumination map in RICD module is involved to adaptively measure the contribution of each neighbor. As a result, IAICD could cope with the complex illumination in challenging nighttime scenarios. 

Finally, considering the nighttime depth estimation \cite{wang2021regularizing,liu2021self,zheng2023steps} is a highly relevant task, we further evaluate our model on it. In summary, our main contributions are as follows: 
\begin{itemize}
    \item For the first time, we extend the conventional depth completion task into challenging nighttime environments to compensate for safe self-driving applications. 
    \item We identify the key challenge of nighttime depth completion as the guidance from color images, where the visibility is low and illumination is complex. Thus we propose RICD and IAICD with learnable differencing centers, which are rather suitable for nighttime scenarios. 
    \item We build two benchmark datasets for the nighttime depth completion task. Extensive experiments indicate the effectiveness of our method, reaching the state of the art. 
\end{itemize}

\section{Related Work}
\textbf{Monocular depth perception at night.} 
Monocular depth perception mainly consists of depth \emph{estimation} \cite{zhou2017unsupervised} and \emph{completion} \cite{Uhrig2017THREEDV}. Up to now, numerous depth estimation methods have been developed in both supervised \cite{liu2023va,piccinelli2023idisc} and self-supervised \cite{zhou2017unsupervised,yan2021channel} ways for daytime scenarios. Recently, some depth estimation approaches \cite{vankadari2020unsupervised,spencer2020defeat,wang2021regularizing,liu2021self,zheng2023steps} focus on nighttime conditions. Specifically, ADDS \cite{liu2021self} proposes a domain-separated network to tackle the large day-night domain shift and illumination variation. RNW \cite{wang2021regularizing} introduces prior regularization and consistent image enhancement for stable training and brightness consistency, respectively. Further, to handle the challenges in underexposed and overexposed regions, STEPS \cite{zheng2023steps} presents a new method that jointly learns a nighttime image enhancer and a depth estimator with uncertain pixel masking strategy and bridge-shaped curve. \emph{For depth completion}, the majority of related works are applied in daytime scenarios, employing either supervised \cite{eldesokey2019confidence,park2020nonlocal,yan2022rignet,lin2022dynamic,Zhang2023CompletionFormer} or self-supervised \cite{wong2021unsupervised,yan2022desnet} manners. For example, RigNet \cite{yan2022rignet} explores a repetitive design in the image guided network branch to acquire clear guidance and structure. CFormer \cite{Zhang2023CompletionFormer} couples convolution and vision transformer to leverage their local and global contexts. However, most of these methods perform poorly at night. 
Considering the challenging nighttime environment is a vital component of self-driving, we attempt to develop a basic framework for nighttime depth completion task to compensate for self-driving applications. 

\textbf{Differencing convolution.} 
Vanilla convolution is commonly used to extract basic visual features in deep learning networks, but it is not very effective when processing scenes with varying illumination. Inspired by Local Binary Pattern \cite{boulkenafet2015face} that is robust to illumination variation, CDC \cite{yu2020searching} first introduces central differencing convolution to aggregate both intensity and gradient information. After that, lots of modified operators are presented in various vision tasks \cite{yu2020fas,su2021pixel,cao2023deep,tan2022semantic}. For example, C-CDC \cite{yu2021dual} extends CDC into dual-cross central differencing convolution via horizontal, vertical, and diagonal decomposition for face anti-spoofing. Further, PDC \cite{su2021pixel} proposes pixel differencing convolution to enhance gradient information for edge detection. Besides, SDN \cite{tan2022semantic} introduces semantic similarity to build semantic differencing convolution for semantic segmentation. However, their fixed differencing centers are not robust and reasonable enough if they contain noise or lie on terminatorn. Differently, our goal is to design learnable differencing centers that are affinitive to neighbor and illumination distribution for safe self-driving at night. 

\textbf{Low-light image enhancement.} 
Low-light images 
often suffer from severe noise, low brightness, low contrast, and color deviation \cite{li2021learning,zhang2023fast}. Thus, plenty of supervised \cite{zhang2021beyond,zhang2021better} and self-supervised \cite{guo2020zero,li2021learning,ma2022toward} methods are presented to restore the details. For example, the well-known histogram equalization (HE) \cite{pizer1987adaptive} is a classic algorithm that strengthens the global contrast. 
However, the accuracy of HE-based approaches would degrade while the background noise contrast increases. As an alternative, Retinex-based methods \cite{zhang2021beyond,liu2021retinex} perform better, with the assumption that low-light image can be decomposed into illumination and reflectance. Recently, SCI \cite{ma2022toward} designs a self-supervised cascaded illumination estimation framework with extremely lightweight parameters. Nevertheless, these methods suffer from either low accuracy or poor robustness in challenging driving scenarios. Different from them, we employ recurrent differencing between paired convolutions to predict reasonable illumination.


\begin{figure*}[t]
\centering
\includegraphics[width=1.82\columnwidth]{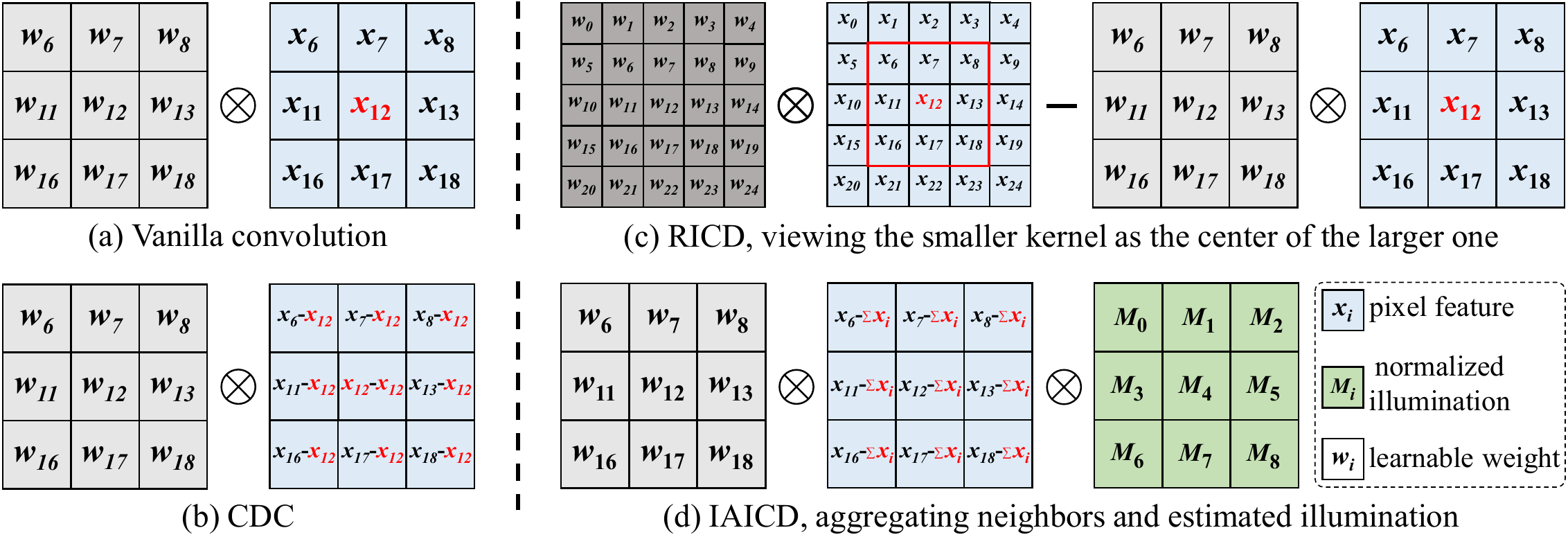}\\
\vspace{-3pt}
\caption{Comparison of vanilla convolution, central differencing convolution (CDC) \cite{yu2020searching}, our recurrent inter-convolution differencing (RICD), and our illumination-affinitive intra-convolution differencing (IAICD). For simplicity, we use $\omega_6$, $\omega_7$, $\cdots$, $\omega_{18}$ to denote a $3\times 3$ convolution kernel.}\label{fig_diff_conv}
\vspace{-3pt}
\end{figure*}

\section{Method}

\subsection{Prior Knowledge}\label{sec_prior}
\textbf{Self-calibrated illumination (SCI).} According to the Retinex theory \cite{land1977retinex}, the relation between the low-light image $\boldsymbol x$ and enhanced image $\boldsymbol x'$ is formulated as: 
\begin{equation}\label{eq_retinex}
\boldsymbol x'=\boldsymbol x \oslash \boldsymbol m,
\end{equation}
where $\boldsymbol m\in (0, 1]$ is the estimated illumination map, and $\oslash$ denotes pixel-wise division. On this basis, the lightweight self-supervised SCI \cite{ma2022toward} designs 
a self-calibrated module and an illumination estimator. Given the low-light input $\boldsymbol x$, this estimator predicts the illumination map $\boldsymbol m$ via several $3\times 3$ convolutions. The enhancement loss includes fidelity and smoothness terms, defined as: 
\begin{equation}\label{eq_lf_ls}
\begin{split}
&{\mathcal{L}_{f}}=\frac{1}{n}\sum_{i=1}^{n}{{\left( \boldsymbol m_i-\boldsymbol x_i \right)}^{2}},\\
&{\mathcal{L}_{s}}=\frac{1}{n}\sum\limits_{i=1}^{n}{\sum\limits_{j\in \mathcal{N}\left( i \right)}{{{\mathcal{G}}_{i,j}}\left| {{\boldsymbol m}_{i}}-{{\boldsymbol m}_{j}} \right|}}, 
\end{split}
\end{equation}
where $\mathcal{G}_{i,j}$ is the weight of a gaussian kernel, and $\mathcal{N}\left( i \right)$ is a window centered at $i$ with $5\times 5$ adjacent pixels. $\mathcal{L}_{f}$ measures the similarity of $\boldsymbol m$ and $\boldsymbol x$ while $\mathcal{L}_{s}$ regularizes the consistency of $\boldsymbol m$ itself. $n$ denotes the number of valid pixels. 

\textbf{Vanilla convolution.} We denote the frequently used 2D spatial convolution as vanilla convolution. 
For simplicity, here we describe the convolution operator in 2D while ignoring the channel dimension. 
Given the input $\boldsymbol x$, the new output feature $\boldsymbol y$ produced by vanilla convolution is represented as: 
\begin{equation}\label{eq_conv}
\boldsymbol y_{p_0}=\sum_{p_n\in \mathcal{R}}{\boldsymbol \omega_{p_n} \cdot \boldsymbol x_{p_0 + p_n}},
\end{equation}
where $\mathcal{R}$ is the local receptive field region sampled from $\boldsymbol x$. $\boldsymbol \omega_{p_0}$ is the convolution weight in current location, whilst $p_n$ enumerates the locations in $\mathcal{R}$. Fig. \ref{fig_diff_conv}(a) is a $3\times 3$ kernel case. 

\textbf{Central differencing convolution.} Based on vanilla convolution, \cite{yu2020searching} designs central differencing convolution (CDC), where every pixel of $\boldsymbol x$ in $\mathcal{R}$ subtracts its center pixel $\boldsymbol x_{p_0}$:
\begin{equation}\label{eq_cdc}
\boldsymbol y_{p_0}=\sum_{p_n\in \mathcal{R}}{\boldsymbol \omega_{p_n} \cdot  \left( \boldsymbol x_{p_0 + p_n} - \boldsymbol x_{p_0} \right )}.
\end{equation}
Fig. \ref{fig_diff_conv}(b) illustrates the process of the above equation. Further, by combining Eqs. \ref{eq_conv} and \ref{eq_cdc}, it yields the trade-off contribution of vanilla convolution and CDC, which is defined as: 

\begin{equation}\label{eq_cdc_final}
\begin{split}
\boldsymbol y_{p_0}&=\theta \cdot \sum_{p_n\in \mathcal{R}}{\boldsymbol \omega_{p_n} \cdot  \left( \boldsymbol x_{p_0 + p_n} - \boldsymbol x_{p_0} \right )}\\
&+ \left( 1 - \theta \right ) \cdot \sum_{p_n\in \mathcal{R}}{\boldsymbol \omega_{p_n} \cdot  \boldsymbol x_{p_0 + p_n}}\\
&\Rightarrow \sum_{p_n\in \mathcal{R}}{\boldsymbol \omega_{p_n} \cdot \boldsymbol x_{p_0 + p_n}} + ( -\boldsymbol x_{p_0} \cdot \sum_{p_n\in \mathcal{R}}{\boldsymbol \omega_{p_n}} ), 
\end{split}
\end{equation}
where the first term is vanilla convolution and the second is central differencing term. 

\subsection{Recurrent Inter-Convolution Differencing}\label{sec_RICD}
Existing low-light image enhancement methods cannot restore very reasonable output in more challenging self-driving nighttime scenarios. For example in Fig.~\ref{fig_color_cast}, SCI \cite{ma2022toward} suffers from serious color cast. To tackle this issue, in Fig.~\ref{fig_diff_conv}(c) we propose recurrent inter-convolution differencing (RICD). RICD first employs convolution subtraction \cite{shi2022symmetric} between two different-kernel vanilla convolutions to highlight the uncertainty of different lighting areas. Then it converts the uncertainty into illumination via recurrent convolution differencing. Suppose that $\mathcal{R}$ is the larger local receptive field region while $\mathcal{\bar{R}}$ is the smaller. $\mathcal{R}$ and $\mathcal{\bar{R}}$ have the same current location $p_0$. As a result, one step of RICD can be formulated as: 
\begin{equation}\label{eq_ricd}
\boldsymbol y_{p_0}=\sum_{p_n\in \mathcal{R}}{\boldsymbol \omega_{p_n} \cdot  \boldsymbol x_{p_0 + p_n}} - \sum_{{\bar{p}_n}\in \mathcal{\bar{R}}}{\boldsymbol \omega_{\bar{p}_n} \cdot  \boldsymbol x_{p_0 + \bar{p}_n}}.
\end{equation}

One novel aspect of RICD is that it converts the uncertainty distribution into illumination estimation. Besides, it introduces a new perspective that identifies the feature of the smaller-kernel convolution as the center of the feature of the larger-kernel convolution. The differencing center is dynamically learned from its local environment. These characteristics contribute to valid illumination prediction. Consequently, according to Eq.~\ref{eq_retinex}, RICD can restore robust enhanced images. 

\begin{figure*}[t]
\centering
\includegraphics[width=1.6\columnwidth]{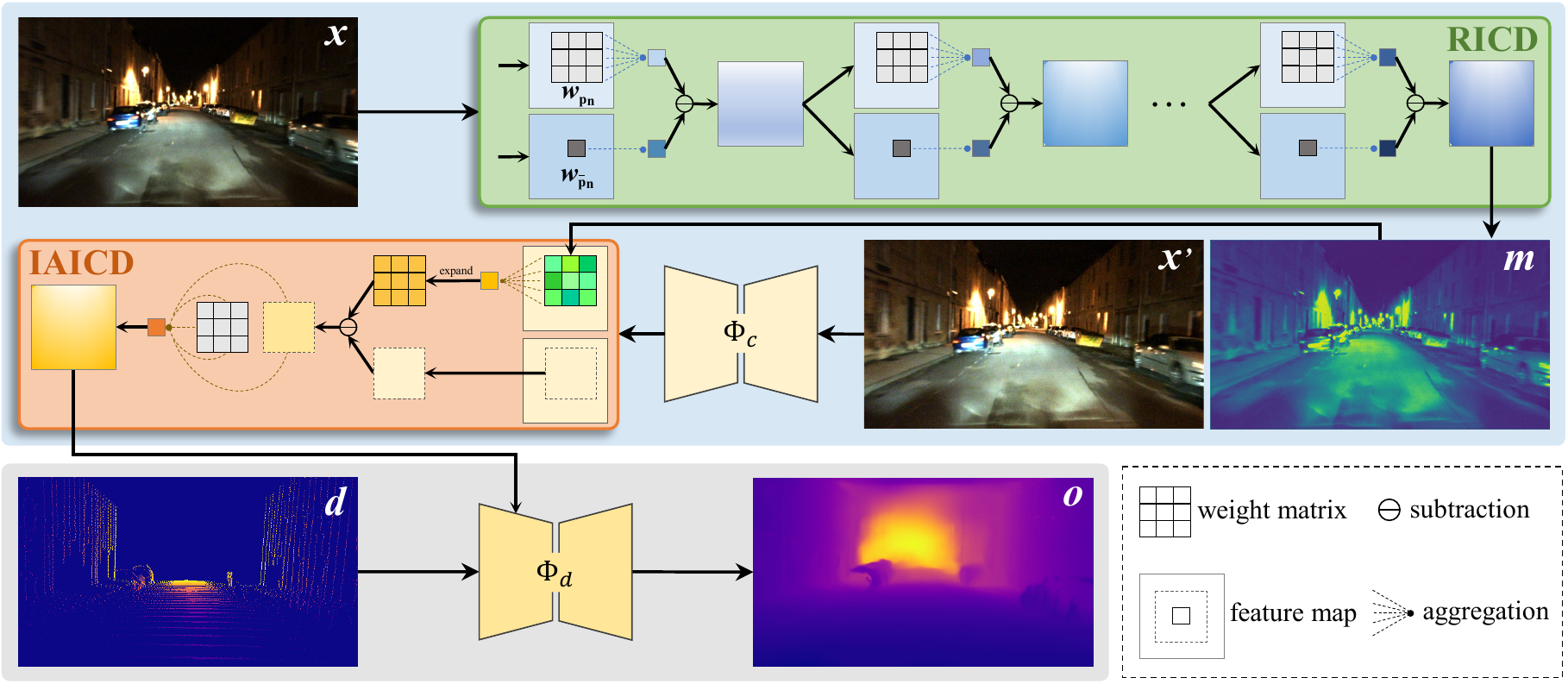}\\
\vspace{-5pt}
\caption{Learnable differencing center network (LDCNet). The low-light input is first fed into RICD to predict credible image and reasonable illumination, based on both of which IAICD is then conducted to alleviate the negative influence of varying illumination.
}\label{fig_pipeline}
\vspace{-3pt}
\end{figure*}

\subsection{Illumination-Affinitive Intra-Convolution Differencing}\label{sec_IAICD}
Although RICD enhances the visibility of nighttime images, the relative light intensity caused by varying illumination is still much more complex than in daytime images. To handle this problem, in Fig.~\ref{fig_diff_conv}(d) we present illumination-affinitive intra-convolution differencing (IAICD). Different from CDC \cite{yu2020searching} whose center is typically fixed, IAICD first aggregates its differencing center adaptively from all neighboring pixels. After yielding the differencing matrix between neighbors and the center, IAICD reweights the matrix via $\mathcal{M}$, which is a channel-wise ($c$) normalization of the illumination map $\boldsymbol m$, \emph{i.e.}, ${{\mathcal{M}}^{c}}={{{\boldsymbol m}^{c}}}/{\sum\limits_{v=1}^{c}{\left| {{\boldsymbol m}^{v}} \right|}}$, then yielding: 
\begin{equation}\label{eq_iaicd}
\boldsymbol y_{p_0}=\sum_{p_n\in \mathcal{R}}{\boldsymbol \omega_{p_n} \cdot ( \boldsymbol x_{p_0 + p_n} - \sum_{p_n\in \mathcal{R}}{\mathcal{M}_{p_n} \cdot \boldsymbol x_{p_n}} )}.
\end{equation}

Compared with CDC, the differencing center predicted by IAICD is robust. For one thing, when the center $\boldsymbol x_{p_0}$ contains noise, CDC would introduce abnormal differencing information whilst IAICD could ignore $\boldsymbol x_{p_0}$ or reduce its negative effect by distributing very small weight. For another, when the center $\boldsymbol x_{p_0}$ lies on terminatorn areas, the fixed $\boldsymbol x_{p_0}$ is no longer appropriate as the differencing center, because its light intensity differs significantly from the neighbors'. As an alternative, we integrate the corresponding illumination map to adjust the weight of each neighboring pixel. 

While the illumination map is a all-ones matrix, and the weight of $p_n (n\neq 0)$ is equal to zero, IAICD will degenerate into CDC. That is to say, IAICD in Eq.~\ref{eq_iaicd} is a generalized version of CDC in Eq.~\ref{eq_cdc}.

\subsection{Learnable Differencing Center Network}\label{sec_LDCNet}
\textbf{Architecture.} The pipeline of our learnable differencing center network (LDCNet) is illustrated in Fig.~\ref{fig_pipeline}. Overall, LDCNet consists of an image guidance branch and a depth prediction branch. In the image guidance branch, the low-light image $\boldsymbol x$ is first fed into RICD, generating the enhanced image $\boldsymbol x'$ and the illumination map $\boldsymbol m$. Next, a simple Unet-like subnetwork $\Phi_c$, composed of five layers with resolutions 1/1, 1/2, 1/4, 1/8, and 1/16, is conducted to encode $\boldsymbol x'$. Together with $\boldsymbol m$, then the features of each layer are input into IAICD. In the depth prediction branch, the sparse depth $\boldsymbol d$ is encoded by a similar subnetwork $\Phi_d$. Meanwhile, the output features of IAICD are resolution-wisely leveraged to guide the dense depth prediction in $\Phi_d$, yielding the final depth output $\boldsymbol o$.

\textbf{Loss Function.} Following previous depth completion methods \cite{tang2020learning,yan2022rignet}, we employ $\mathcal{L}_2$ loss to supervise the output $\boldsymbol o$ by using groundtruth depth $\boldsymbol D$: 
\begin{equation}\label{eq_l2}
{\mathcal{L}_{2}}=\frac{1}{n}\sum_{i=1}^{n}{{\left( \boldsymbol D_i-\boldsymbol o_i \right)}^{2}}.
\end{equation}
Finally, we jointly train the low-light image enhancement subnetwork and depth prediction subnetwork by combining Eqs.~\ref{eq_lf_ls} and~\ref{eq_l2}, obtaining the total loss function: 
\begin{equation}\label{eq_final_loss}
\mathcal{L}_{total}=\mathcal{L}_2 + \alpha \mathcal{L}_f + \beta \mathcal{L}_s,
\end{equation}
where $\alpha$ and $\beta$ are hyper-parameters, which are set to 0.15 and 0.3 as the default, respectively. 

\section{Experiments}

\subsection{Datasets and Implementation Details.} 

\textbf{RobotCar-Night-DC.} Oxford RobotCar \cite{maddern20171} is a large-scale dataset that captures various weather and traffic conditions along a route in central Oxford. We create RobotCar-Night-DC from the 2014-12-16-18-44-24 sequences by using the left color images of the front stereo-camera (Bumblebee XB3). To generate sparse and groundtruth depth maps, we employ the official toolbox to process the data from the front LMS laser and INS sensors. Following KITTI benchmark \cite{Uhrig2017THREEDV}, we use the current frame for sparse depth generation and multiple frames for groundtruth depth creation. The densities of the valid pixels of sparse depth and groundtruth depth are about $4\%$ and $16\%$, respectively. Then we crop and resize these data to $576\times 320$ to remove the car-hood and enable efficient training. As a result, the RobotCar-Night-DC dataset contains $10,290$ RGB-D pairs for training and $411$ for testing. 

\textbf{CARLA-Night-DC.} CARLA-EPE \cite{zheng2023steps} is a synthetic dataset for nighttime depth estimation task, generated by CARLA simulator \cite{dosovitskiy2017carla} and EPE network \cite{richter2022enhancing}. The groundtruth depth in CARLA-EPE is almost fully dense, which is unrealistic for LiDAR-based self-driving systems where the depth density is around $7\%$ \cite{yan2022rignet}. Hence, based on the synthetic dataset we create CARLA-Night-DC for the proposed nighttime depth completion task, by transferring the sparse LiDAR pattern of KITTI \cite{Uhrig2017THREEDV} to CARLA-EPE. Hence, CARLA-Night-DC is composed of $7,532$ RGB-D pairs in total, of which $7,000$ for training and $532$ for testing. 

\textbf{Implementation Details}. We implement LDCNet using Pytorch on a single RTX 3090 GPU. We train it for $20$ epochs with the Adam optimizer, the momentum $\beta_{1}=0.9$, $\beta_{2}=0.999$, and weight decay $1 \times {10}^{-6}$. The initial learning rate is $1 \times {10}^{-3}$ that drops by half every $5$ epochs. We use synchronized cross-GPU batch normalization \cite{ioffe2015batch}, resulting in a batch size of $12$. Evaluation metrics are consistent with RNW \cite{wang2021regularizing} and KITTI \cite{Uhrig2017THREEDV}. \underline{RMSE} is measured \underline{in meters}. 

\begin{table*}[t]
\centering
\renewcommand\arraystretch{1.15}
\scriptsize
\caption{Results of \emph{nighttime depth estimation} on RobotCar \cite{zheng2023steps} and CARLA \cite{zheng2023steps} benchmarks.}\label{tab_estimation}
\vspace{-7pt}
\resizebox{0.865\textwidth}{!}{
\begin{tabular}{l|cccc|ccc}
\hline
\cellcolor[RGB]{255,217,102}Method    & \cellcolor[RGB]{244,176,132}Abs Rel $\downarrow$ & \cellcolor[RGB]{244,176,132}Sq Rel $\downarrow$ & \cellcolor[RGB]{244,176,132}RMSE $\downarrow$ & \cellcolor[RGB]{244,176,132}RMSE log $\downarrow$ & \cellcolor[RGB]{155,194,230}${\delta }_1$ $\uparrow$ & \cellcolor[RGB]{155,194,230}${\delta }_2$ $\uparrow$ & \cellcolor[RGB]{155,194,230}${\delta }_3$ $\uparrow$ \\ \hline
\multicolumn{8}{c}{\cellcolor[RGB]{198,224,180}RobotCar}  \\ \hline
MD2 \cite{godard2019digging}     & 0.580 & 21.446 & 12.771 & 0.521 & 0.552 & 0.840 & 0.920 \\
DeFeatNet \cite{spencer2020defeat} & 0.334 & 4.589 & 8.606 & 0.358 & 0.586 & 0.827 & 0.911  \\
ADFA \cite{vankadari2020unsupervised}   & 0.233 & 3.783 & 10.089 & 0.319 & 0.668 & 0.884 & 0.924 \\
ADDS \cite{liu2021self}                 & 0.231 & 2.674 & 8.800 & 0.268 & 0.620 & 0.892 & 0.956 \\
RNW \cite{wang2021regularizing}         & 0.185 & 1.894 & 7.319 & 0.246 & 0.735 & 0.910 & 0.965 \\
WSGD \cite{vankadari2023sun} & 0.174 & 1.637 & \textbf{6.302} & 0.245 & 0.754 & 0.915 & 0.964 \\
STEPS \cite{zheng2023steps}             & \underline{0.170} & \underline{1.686} & 6.797 & \underline{0.234} & \underline{0.758} & \underline{0.923} & \underline{0.968} \\
LDCNet (ours)                           & \textbf{0.161} & \textbf{1.555} & \underline{6.725} & \textbf{0.228} & \textbf{0.781} & \textbf{0.925} & \textbf{0.970} \\ \hline
\multicolumn{8}{c}{\cellcolor[RGB]{198,224,180}CARLA} \\ \hline
MD2 \cite{godard2019digging}     & 0.555 & 6.974 & 9.761 & 0.549 & 0.329 & 0.594 & 0.781 \\
RNW \cite{wang2021regularizing}         & 0.485 & 6.308 & 8.557 & 0.483 & 0.404 & 0.703 & 0.849 \\
STEPS \cite{zheng2023steps}     & \underline{0.481} & \underline{6.267} & \underline{8.519} & \underline{0.480} & \underline{0.406} & \underline{0.706} & \underline{0.851} \\
LDCNet (ours)                           & \textbf{0.463} & \textbf{6.026} & \textbf{8.472} & \textbf{0.469} & \textbf{0.420} & \textbf{0.721} & \textbf{0.863} \\ \hline
\end{tabular}
}
\vspace{-7pt}
\end{table*}

\begin{figure*}[t]
\centering
\includegraphics[width=1.755\columnwidth]{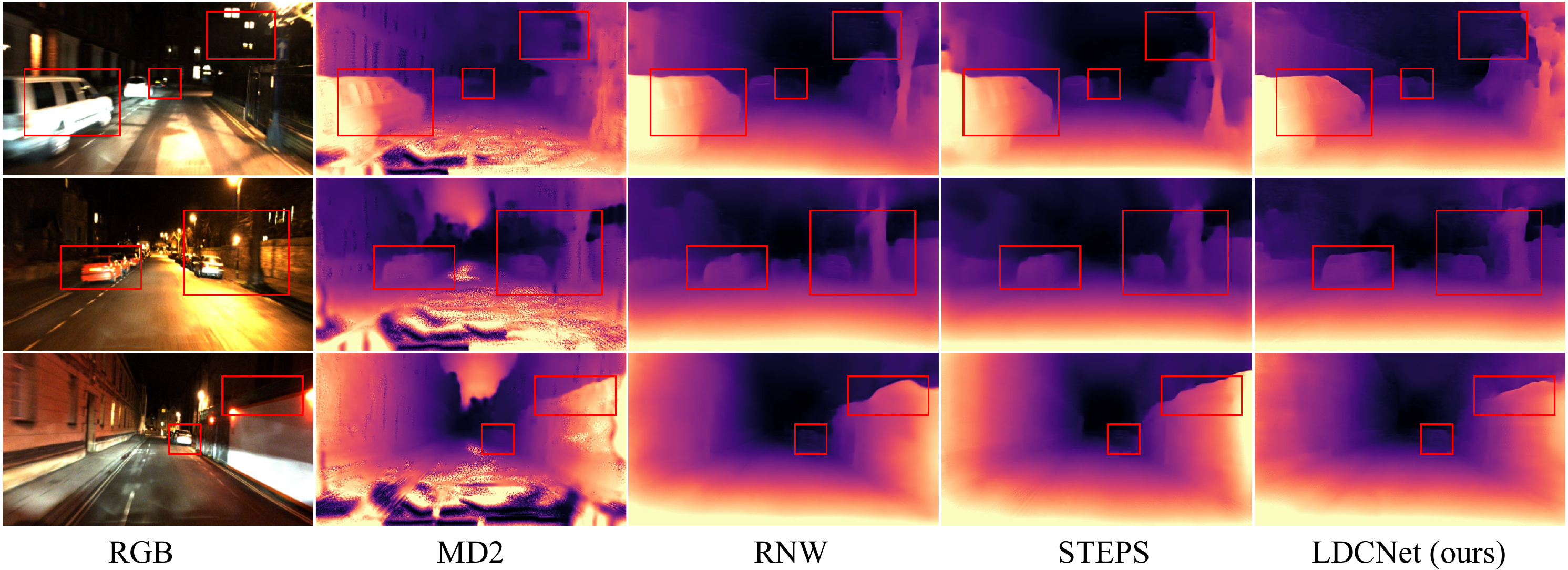}\\
\vspace{-7pt}
\caption{Visual comparison of nighttime depth estimation on RobotCar-Night-DC dataset.}\label{fig_robotcar_vis}
\vspace{-7pt}
\end{figure*}

\subsection{Results}

\textbf{Nighttime depth estimation.} We compare LDCNet with \emph{nighttime} state-of-the-art methods, including MD2 \cite{godard2019digging}, DeFeatNet \cite{spencer2020defeat}, ADFA \cite{vankadari2020unsupervised}, ADDS \cite{liu2021self}, RNW \cite{wang2021regularizing}, WSGD \cite{vankadari2023sun}, and STEPS \cite{zheng2023steps}. Based on STEPS, we embed our RICD and IAICD into its image enhancement branch and depth estimation branch, respectively. From Tab.~\ref{tab_estimation} we can observe that LDCNet almost achieves the lowest errors and the highest accuracy. On RobotCar dataset, LDCNet is superior to the second best STEPS in all aspects. Furthermore, LDCNet surpasses the well-known MD2 by large margins. For example, the RMSE of MD2 is reduced from $12.771m$ to $6.725m$, almost $47\%$ improvement, whilst the accuracy $\delta_1$ acquires an increase of $22.9$ percentage points. On CARLA dataset, LDCNet also performs better than other three approaches. In addition, we compare these methods on RobotCar and CARLA and observe that they perform worse on CARLA. This can be attributed to the darker color images and the larger depth ranges of CARLA. Finally, the visual results in Fig.~\ref{fig_robotcar_vis} shows that LDCNet can predict more accurate depth maps with more complete and sharper edges, which further verify the superiority and effectiveness of LDCNet.

\begin{table*}[t]
\centering
\renewcommand\arraystretch{1.15}
\caption{Results of \emph{nighttime depth completion} on RobotCar-Night-DC and CARLA-Night-DC.}
\label{tab_completion}
\vspace{-7pt}
\resizebox{0.912\textwidth}{!}{
\begin{tabular}{l|cccc|cccc}
\hline
\cellcolor[RGB]{255,217,102}Method    & \cellcolor[RGB]{244,176,132}RMSE $\downarrow$ & \cellcolor[RGB]{244,176,132}MAE $\downarrow$ & \cellcolor[RGB]{244,176,132}iRMSE $\downarrow$ & \cellcolor[RGB]{244,176,132}iMAE $\downarrow$ & \cellcolor[RGB]{244,176,132}RMSE $\downarrow$ & \cellcolor[RGB]{244,176,132}MAE $\downarrow$ & \cellcolor[RGB]{244,176,132}iRMSE $\downarrow$ & \cellcolor[RGB]{244,176,132}iMAE $\downarrow$ \\ \hline
\multicolumn{9}{c}{\cellcolor[RGB]{198,224,180}  \quad  \quad  \quad \quad \quad \quad  \quad RobotCar-Night-DC \quad \quad \quad \quad \quad \quad \quad \quad  \quad \quad \quad CARLA-Night-DC} \\ \hline
NCNN \cite{eldesokey2019confidence} & 6.397  & 5.341  & 0.0532  & 0.0389  & 34.956  & 23.246 
 & 0.2785  & 0.1204 \\
pNCNN \cite{eldesokey2020uncertainty} & 5.879  & 4.663  & 0.0449  & 0.0331 
 & 43.929  & 32.712  & 0.2914  & 0.1321 \\
S2D \cite{ma2018self}     & 5.251 & 6.115 & 0.8832 & 0.0571 & 13.472 & 3.534 & 0.0577 & 0.0204 \\
NLSPN \cite{park2020nonlocal}   & 4.586 & 2.994 & 0.2536 & 0.0283 & 36.008 & 19.760 & 0.0581 & 0.0217 \\
FusionNet \cite{vangansbeke2019} & 1.133  & 0.453  & 0.0067  &  0.0027  & 35.849  & 17.043 
 & 0.0580  & 0.0193 \\
GuideNet \cite{tang2020learning}     & 1.321 & 0.681 & 0.0074 & 0.0038 & 8.019 & 2.710 & \underline{0.0574} & \underline{0.0160} \\
RigNet \cite{yan2022rignet}    & 1.285 & 0.654 & 0.0073 & 0.0036 & \underline{7.675} & \underline{2.259} & 0.0577 & 0.0165\\
CFormer \cite{Zhang2023CompletionFormer}   & \underline{1.183} & \underline{0.473} & \underline{0.0064} & \underline{0.0024} & 33.669 & 18.391 & 0.0578 & 0.0211 \\
LDCNet (ours)        & \textbf{1.170} & \textbf{0.466} & \textbf{0.0059} & \textbf{0.0023} & \textbf{7.214} & \textbf{2.014} & \textbf{0.0546} & \textbf{0.0156} \\ \hline
\end{tabular}
}
\vspace{-7pt}
\end{table*}

\begin{figure*}[t]
\centering
\includegraphics[width=1.8\columnwidth]{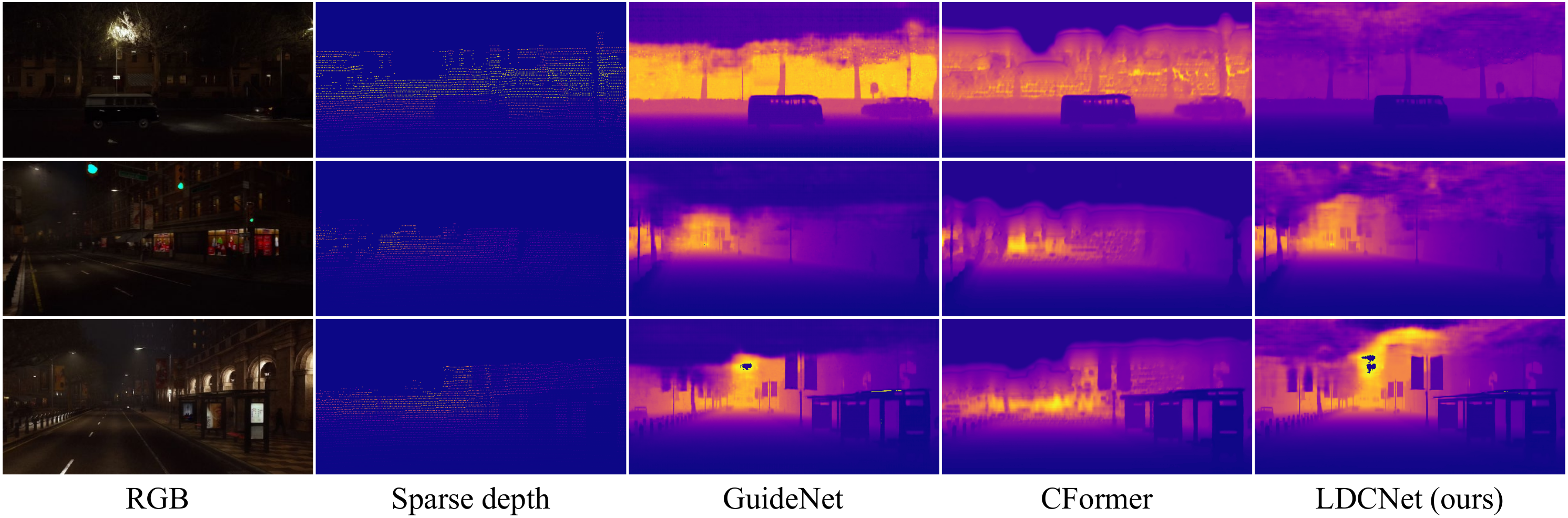}\\
\vspace{-7pt}
\caption{Visual comparison of nighttime depth completion on CARLA-Night-DC dataset.}\label{fig_carla_vis}
\vspace{-7pt}
\end{figure*}

\textbf{Nighttime depth completion.} For fair comparison, we \textbf{retrain} existing state-of-the-art \emph{daytime} depth completion approaches in nighttime scenarios, including FusionNet \cite{vangansbeke2019}, NCNN \cite{eldesokey2019confidence}, pNCNN \cite{eldesokey2020uncertainty}, S2D \cite{ma2018self}, NLSPN \cite{park2020nonlocal}, GuideNet \cite{tang2020learning}, RigNet \cite{yan2022rignet}, and CFormer \cite{Zhang2023CompletionFormer}. 
The quantitative results are reported in Tab.~\ref{tab_completion}. Overall, we discover that LDCNet achieves the best performance on the two nighttime depth perception benchmarks. Specifically, on RobotCar-Night-DC dataset, LDCNet is comprehensively superior to other methods. For instance, LDCNet reduces the MAE by $28.7\%$ over the third best RigNet. Compared with the second best CFormer, which requires 5 days for training on a single 3090 GPU, LDCNet still achieves slightly better results with 20-hour training cost. On CARLA-Night-DC dataset, the challenging darker environment and greater distance result in poor performance of these methods. For example, the RMSE is at least $6m$ larger than that on RobotCar-Night-DC. Additionally, we notice that NCNN, pNCNN, NLSPN, FusionNet, and CFormer, 
all of which estimate confidence map to reweight depth, suffer from quite large RMSE and MAE. We analyse that the very low-light color images make it rather difficult to predict credible confidence distribution, resulting in unstable depth refinement. At last, from Fig.~\ref{fig_carla_vis} we discover that LDCNet succeeds in recovering object depth more accurately, such as the cars, bus shelters, and buildings in the foreground, and the trees, light poles, and billboards in the background.

\subsection{Ablation Study}
For efficient ablation on RobotCar-Night-DC, we halve the size of the two subnetworks in LDCNet by setting the stride of the first-layer convolution to 2. 

\textbf{LDCNet.} As reported in Tab.~\ref{tab_ab_ldcnet}, the baseline LDCNet-i 
first removes RICD and IAICD modules. Then, as an alternative to IAICD, LDCNet-i incorporates the guidance module proposed in GuideNet \cite{tang2020learning}. When implementing our RICD design (LDCNet-ii), we discover that the two evaluation metrics are consistently improved, \emph{i.e.}, RMSE is reduced by $104mm$ and MAE by $129mm$. Similarly, the individual IAICD (LDCNet-iii) contributes to larger performance improvement, severally reducing RMSE and MAE by $117mm$ and $148mm$. Finally, to combine the best of two worlds, LDCNet-iv embeds RICD and IAICD simultaneously into the baseline. As a result, LDCNet-iv performs much better than LDCNet-i, significantly exceeding it by $137mm$ in RMSE and $181mm$ in MAE. 

\begin{table}[t]
\centering
\scriptsize
\renewcommand\arraystretch{1.15}
\caption{Ablation on components of LDCNet.}
\label{tab_ab_ldcnet}
\vspace{-7pt}
\resizebox{0.425\textwidth}{!}{
\begin{tabular}{l|cc|cc}
\hline
LDCNet  & RICD        & IAICD      & RMSE    & MAE     \\ \hline
i       &             &            & 1.321   & 0.681   \\
ii      & \checkmark  &            & 1.217   & 0.552   \\
iii     &             & \checkmark & 1.204   & 0.533   \\
iv      & \checkmark  & \checkmark & \textbf{1.184}  & \textbf{0.500}  \\ \hline
\end{tabular}
}
\vspace{-7pt}
\end{table}

\begin{table}[t]
\centering
\scriptsize
\renewcommand\arraystretch{1.15}
\caption{Ablation on diverse-kernel RICD.}
\label{tab_ab_ricd}
\vspace{-7pt}
\resizebox{0.421\textwidth}{!}{
\begin{tabular}{l|cc|cc}
\hline
RICD   \quad \quad  & $k_1$ & $k_2$               & RMSE    & MAE     \\ \hline
i      & $3\times 3$  & $1\times 1$  & 1.295   & 0.640   \\
ii     & $5\times 5$  & $3\times 3$  & 1.217   & 0.552   \\
iii    & $7\times 7$  & $5\times 5$  & \textbf{1.203}   & \textbf{0.531}   \\
iv     & $7\times 7$  & $3\times 3$  & 1.241   & 0.581   \\ \hline
\end{tabular}
}
\vspace{-7pt}
\end{table}

\begin{figure}[t]
\centering
\includegraphics[width=0.98\columnwidth]{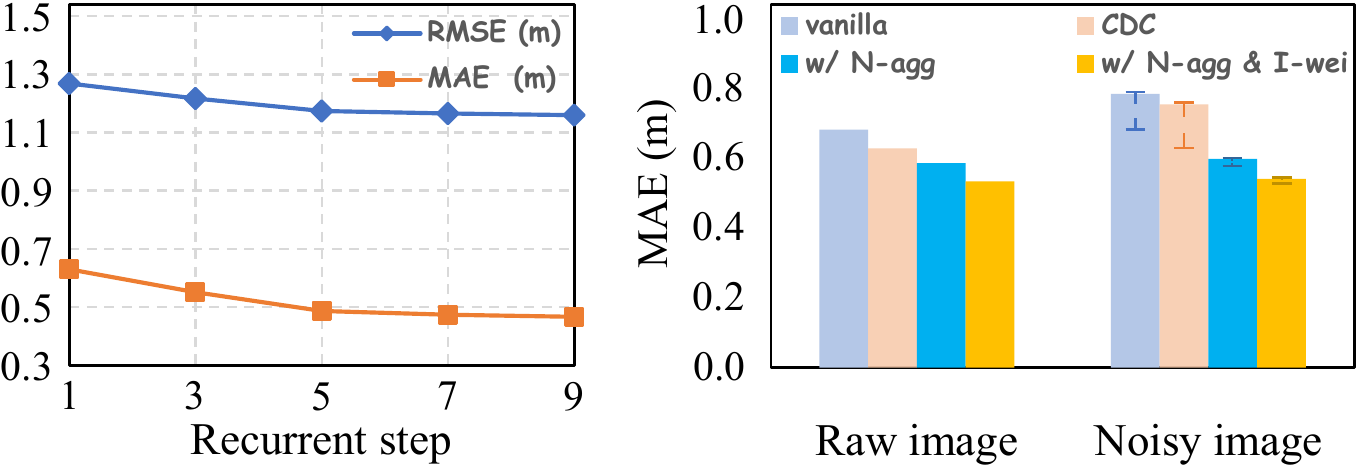}\\
\vspace{-7pt}
\caption{Ablation on RICD (left) and IAICD (right). `N-agg': neighboring aggregation. `I-wei': illumination-affinitive weighting. }\label{fig_ab}
\vspace{-3pt}
\end{figure}
\begin{figure}[t]
\centering
\includegraphics[width=1\columnwidth]{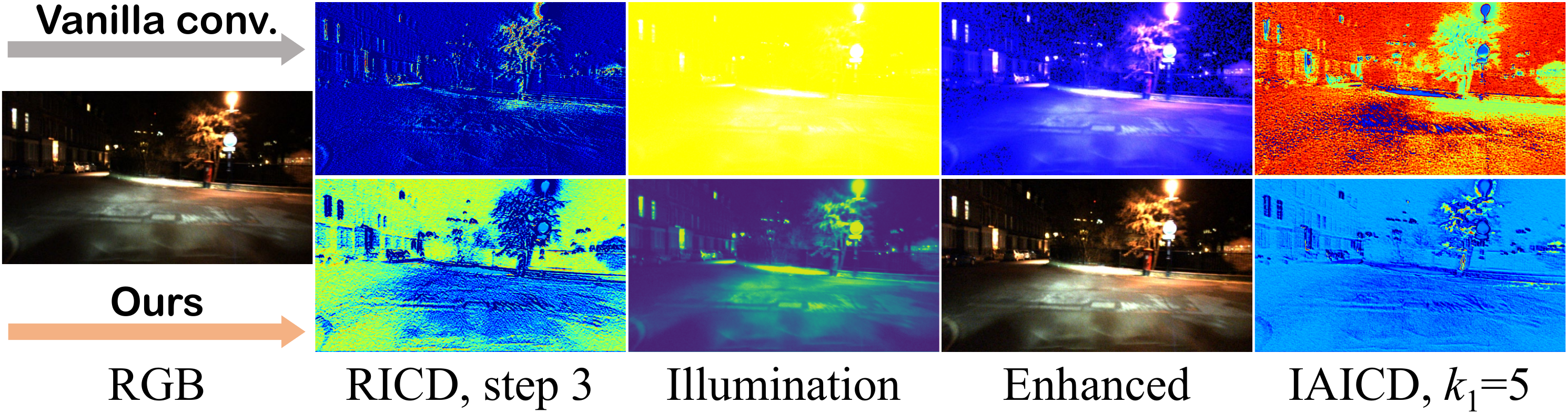}\\
\vspace{-7pt}
\caption{Feature comparison of vanilla convolution and our method.}\label{fig_mid_feat_vis}
\vspace{-4pt}
\end{figure}

\begin{table}[!ht]
\centering
\renewcommand\arraystretch{1.12}
\caption{Results on \href{http://www.cvlibs.net/datasets/kitti/eval_depth.php?benchmark=depth_completion}{KITTI depth completion benchmark}.}
\label{tab_gc_ldcnet}
\vspace{-6pt}
\resizebox{0.489\textwidth}{!}{
\begin{tabular}{l|cccc}
\hline
\cellcolor[RGB]{255,217,102}Method             & \cellcolor[RGB]{244,176,132}RMSE $\downarrow$ & \cellcolor[RGB]{244,176,132}MAE $\downarrow$ & \cellcolor[RGB]{244,176,132}iRMSE $\downarrow$ & \cellcolor[RGB]{244,176,132}iMAE $\downarrow$  \\ \hline
CSPN \cite{2018Learning}          & 1019.64   & 279.46   & 2.93    & 1.15  \\
DesNet \cite{yan2022desnet}       & 938.45    & 266.24   & 2.95    & 1.13  \\
DLiDAR \cite{Qiu_2019_CVPR}       & 758.38    & 226.50   & 2.56    & 1.15  \\
GuideNet \cite{tang2020learning}    & 736.24    & 218.83   & 2.25    & 0.99  \\
RigNet \cite{yan2022rignet}        & 712.66  & \underline{203.25}  & 2.08  & 0.90  \\ 
DySPN \cite{lin2022dynamic}        & \underline{709.12}  & \textbf{192.71}  & \textbf{1.88}  & \textbf{0.82}  \\ 
CFormer \cite{Zhang2023CompletionFormer} & \textbf{708.87}  & 203.45  & \underline{2.01}  & \underline{0.88}  \\ \hline
LDCNet (ours)   & 753.15  & 218.02  & 2.33  & 0.98  \\ \hline
\end{tabular}
}
\vspace{-6pt}
\end{table}

\textbf{RICD.} The basic unit of RICD is the differencing between two convolutions with different kernels. Consequently, we ablate diverse kernel sizes in Tab. \ref{tab_ab_ricd}. Based on LDCNet-i, RICD-i, RICD-ii, and RICD-iii conduct $(k+2)\times (k+2)$ and $k\times k$ convolution differencing. We can find that, as the kernel size increases, the two evaluation metrics decrease gradually. For example, the MAE of $k_2=5$ is $109mm$ superior to that of $k_2=1$. This is due to the learnable differencing center design, which regards the small-kernel-convolution feature as the center of the large-kernel-convolution feature. Such differencing convolutions with larger local receptive fields can predict reliable illumination distribution by aggregating the surrounding light information. Further, RICD-iv increases the kernel size gap from $2$ to $4$. For one thing, it is clear that the $1\times 1$ convolution of RICD-i is not very suitable to be the differencing center because it cannot leverage ambient information. Thus, RICD-iv performs better than RICD-i regardless of the larger gap. For another thing, with the larger size gap, the larger-kernel convolution would introduce redundant light reference over long distances, while the smaller-kernel convolution can only map the light in local regions. Therefore, RICD-iv performs worse than RICD-ii and RICD-iii with smaller size gap. In addition, based on RICD-ii, Fig.~\ref{fig_ab} (left) shows the ablation on RICD with different recurrent steps. We observe that RICD performs better as the step grows. As shown in Fig.~\ref{fig_mid_feat_vis}, RICD can strengthen the representation of relative light intensity, contributing to more precise illumination. Finally, we select RICD-ii and step-3 as the default. 

\textbf{IAICD.} Different from the center differencing convolution (CDC) \cite{yu2020searching} with fixed center, IAICD first aggregates all neighboring pixels and then employs the illumination-affinitive weight to produce its learnable center. Fig.~\ref{fig_ab} (right) shows that both of these two strategies contribute to consistent improvement over vanilla convolution and CDC. Furthermore, to evaluate the robustness of IAICD, we introduce Gaussian noise into raw color images. As can be seen, IAICD still performs better than CDC and achieves very close performance to itself using raw color images. All of these evidences demonstrate the effectiveness and robustness of IAICD.

\subsection{Generalization}
Here we further evaluate the generalization capabilities of our LDCNet on both daytime depth completion \cite{Uhrig2017THREEDV} and low-light image enhancement \cite{zhang2021beyond} tasks. 

\begin{figure}[t]
\centering
\includegraphics[width=0.98\columnwidth]{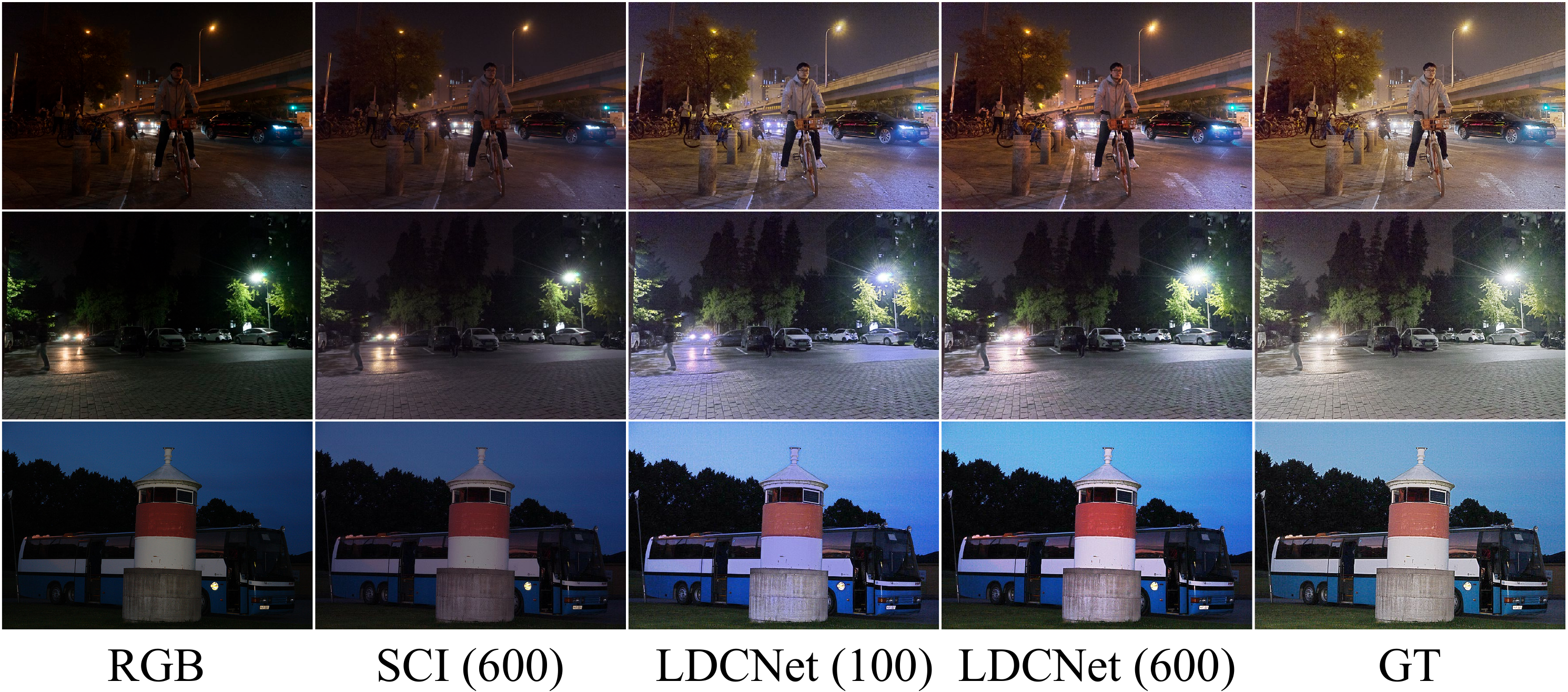}\\
\vspace{-6pt}
\caption{Low-light image enhancement on \href{https://github.com/vis-opt-group/SCI/tree/main/data/difficult}{difficult test split} of SCI \cite{ma2022toward}.}\label{fig_image_enhance_vis}
\vspace{-2pt}
\end{figure}

\begin{table}[t]
\centering
\scriptsize
\renewcommand\arraystretch{1.12}
\caption{Comparison on \href{https://github.com/vis-opt-group/SCI/tree/main/data/difficult}{difficult test split} of SCI.}
\label{tab_enhance_ldcnet}
\vspace{-5pt}
\resizebox{0.482\textwidth}{!}{
\begin{tabular}{l|cc|cc}
\hline
\cellcolor[RGB]{255,217,102}Method  & \cellcolor[RGB]{244,176,132}NIQE $\downarrow$   & \cellcolor[RGB]{244,176,132}DE $\uparrow$   & \cellcolor[RGB]{155,194,230}PSNR $\uparrow$     & \cellcolor[RGB]{155,194,230}SSIM $\uparrow$    \\ \hline
SCI \cite{ma2022toward}  & 3.3510           & 6.3300           & 12.2867           & 0.5034            \\
+ RICD                   & \textbf{3.0816}  & \textbf{7.2192}  & \textbf{21.5286}  & \textbf{0.8992}   \\ \hline
\end{tabular}
}
\vspace{-4pt}
\end{table}

Tab.~\ref{tab_gc_ldcnet} reports the comparison results on KITTI depth completion dataset \cite{Uhrig2017THREEDV}, which is collected during the daytime. We can observe that the performance of current state-of-the-art methods \cite{tang2020learning,yan2022rignet,lin2022dynamic,Zhang2023CompletionFormer} is very similar. For example, the ranking metric RMSE is $730mm$ nearby. Although our LDCNet is specifically designed for nighttime scenarios, it still achieves competitive performance on the daytime benchmark. 

Based on the self-supervised SCI \cite{ma2022toward} that is trained for $600$ epochs, we replace its illumination estimation module with our RICD block. From Tab.~\ref{tab_enhance_ldcnet} we can discover that RICD consistently improves the baseline in both \emph{no-reference} NIQE \cite{wang2013naturalness} \& DE \cite{shannon1948mathematical} and \emph{full-reference} PSNR \& SSIM metrics. Furthermore, Fig.~\ref{fig_image_enhance_vis} demonstrates the superiority of our method again, \emph{i.e.}, higher quality with lower training cost.

\section{Conclusion}
In this paper, we extended the conventional depth completion task into nighttime environments to complement safe self-driving. We identified the key challenge as the guidance from color images with low visibility and complex illumination. As a result, we proposed RICD and IAICD to improve the poor visibility and reduce negative influences of the varying illumination, respectively. RICD could predict explicit global illumination to enhance visibility, where treating the small-kernel convolution as the center of the large-kernel-convolution was a new perspective. IAICD succeeded in alleviating local relative light intensity, in which the differencing center was learned dynamically from the neighboring pixels and illumination maps of RICD. Thus, the center was robust and illumination-affinitive. Finally, extensive experiments on depth perception datasets have verified the effectiveness of LDCNet. 


\bibliographystyle{IEEEtran}
\bibliography{egbib}



\end{document}